\documentclass[letterpaper]{article} 
\usepackage{aaai2026}  
\usepackage{times}  
\usepackage{helvet}  
\usepackage{courier}  
\usepackage[hyphens]{url}  
\usepackage{graphicx} 
\usepackage{natbib}  
\usepackage{caption} 
\usepackage{algorithm}
\usepackage{algorithmic}

\usepackage{amsmath}
\usepackage{amsfonts}
\usepackage{amssymb}
\usepackage{overpic}
\usepackage{multirow}
\usepackage{mathtools}
\usepackage{makecell}
\usepackage{lscape}
\usepackage{newfloat}
\usepackage{listings}
\DeclareCaptionStyle{ruled}{labelfont=normalfont,labelsep=colon,strut=off} 
\floatstyle{ruled}
\newfloat{listing}{tb}{lst}{}
\floatname{listing}{Listing}
\title{Hide and Seek with LLMs: An Adversarial Game for Sneaky Error Generation and Self-Improving Diagnosis}

\author{
Rui Zou\textsuperscript{1},
Mengqi Wei\textsuperscript{2},
Yutao Zhu\textsuperscript{1},
Jirong Wen\textsuperscript{1}\thanks{Corresponding author: jrwen@ruc.edu.cn.},
Xin Zhao\textsuperscript{1},
Jing Chen\textsuperscript{3}
}

\affiliations{
\textsuperscript{1}Gaoling School of Artificial Intelligence, Renmin University of China \\
\textsuperscript{2}Faculty of Artificial Intelligence in Education, Central China Normal University \\
\textsuperscript{3}School of Information and Security Engineering, Zhongnan University of Economics and Law \\
zouruixyz@ruc.edu.cn
}

\usepackage{bibentry}

\begin{document}

\maketitle

\begin{abstract}
    Large Language Models (LLMs) excel in reasoning and generation across domains, but still struggle with identifying and diagnosing complex errors. This stems mainly from training objectives that prioritize correct answers, limiting exposure to and learning from errors. While recent studies have begun to address this by introducing error signals, most rely on shallow, static errors, restricting improvement in deep diagnostic ability. To overcome this, we propose Hide and Seek Game (HSG), a dynamic adversarial framework for error generation and diagnosis, and evaluate it on mathematical problem-solving. HSG involves two adversarial roles: Sneaky, which ``hides'' by generating subtle, deceptive reasoning errors, and Diagnosis, which ``seeks'' to accurately detect them. Through adversarial co-evolution, both error stealth and diagnostic precision are enhanced. Experiments on several math reasoning tasks show that HSG significantly boosts error diagnosis, achieving 16.8\%--31.4\% higher accuracy than baselines like GPT-4o. We also release a challenging dataset of deceptive errors and diagnostic annotations as a benchmark for future research.
\end{abstract}

\section{Introduction}

LLMs have achieved impressive results in tasks such as question answering and text generation \cite{chang2024survey,wang2023document,wei2022chain}, and are widely adopted in domains like education \cite{wang2024large} and healthcare \cite{maity2025large}. However, a persistent challenge is their limited ability to recognize and diagnose errors. Studies have shown that without external feedback, LLMs often fail to proactively identify and correct mistakes in their reasoning \cite{ICLR2024_8b4add8b,tyen2024llms,lin2024criticbench,liang2025mathclean}. This is largely because mainstream training focuses on producing correct answers, giving models little exposure to diverse error patterns and correction processes, and thus restricting their error analysis capabilities.

This lack of diagnostic skill poses significant risks, especially in high-stakes scenarios. For example, an autonomous driving model that cannot detect logical flaws in its planned route may cause catastrophic accidents \cite{shalev2017formal}; in intelligent tutoring, a model unable to identify student misconceptions may undermine personalized guidance \cite{wang2024large}; and in legal document generation, failure to spot hidden factual contradictions can lead to serious disputes \cite{bewersdorff2023assessing}. Enhancing LLMs’ error diagnosis is therefore vital for building trustworthy and robust AI systems.

Moreover, error recognition is a key driver for continual model improvement. Advanced approaches like Direct Preference Optimization (DPO) \cite{rafailov2023direct} and Reinforcement Learning from Human Feedback (RLHF) \cite{ouyang2022training} improve model judgment by reinforcing correct outputs and penalizing mistakes. Recent work \cite{tyen2024llms} also shows that improving error localization significantly boosts reasoning output quality.

\begin{figure}[t]
    \centering
    \includegraphics[width=0.42\textwidth]{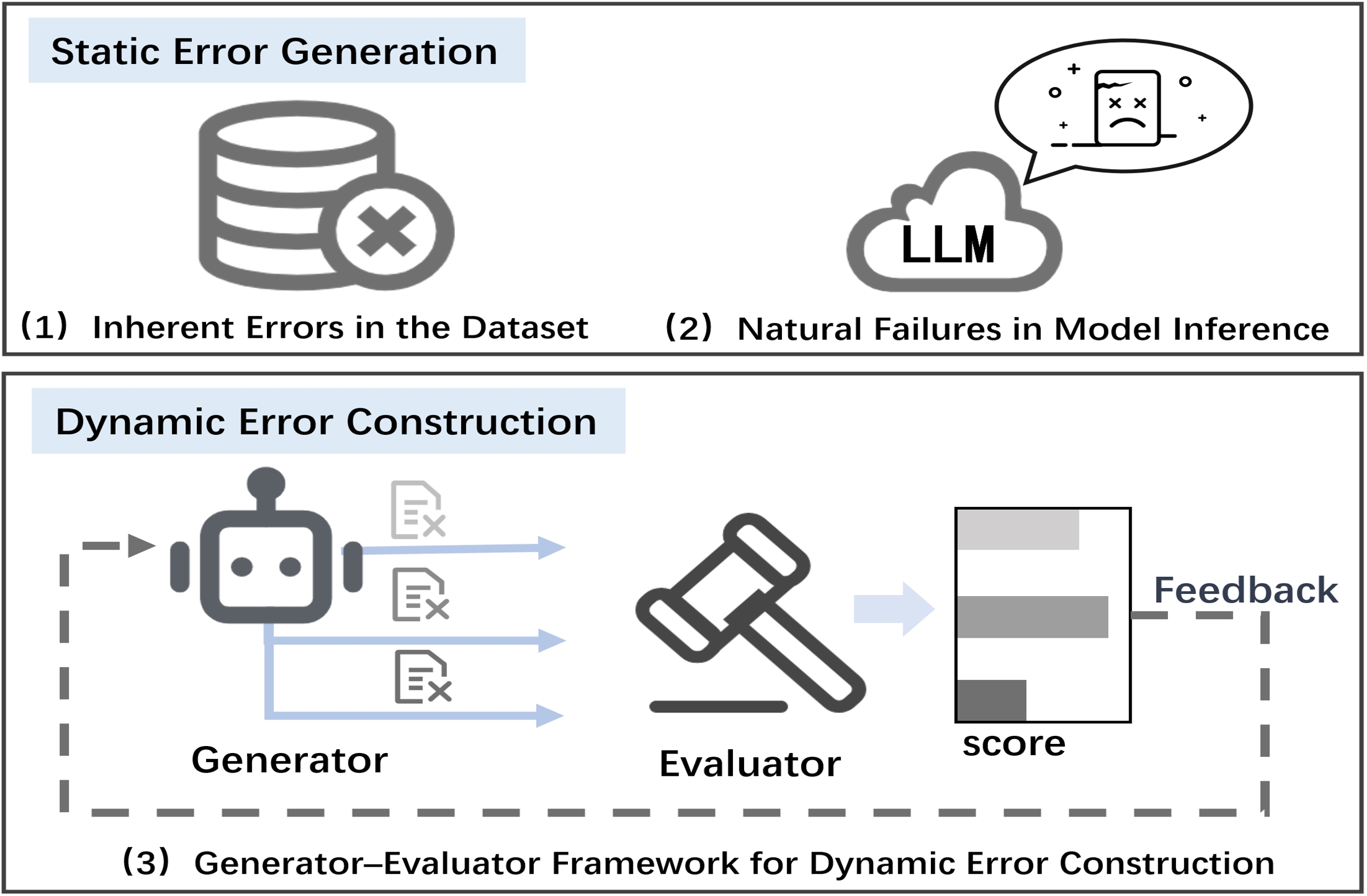}
    \caption{Comparison of three error utilization strategies.}
    \label{fig:sec1_compare}
\end{figure}

Given the importance of error diagnosis for both reliability and continual learning, a key question arises: how can we enhance LLMs’ error recognition and localization? A promising approach is inspired by human teaching—where teachers design hard-to-diagnose problems to uncover students’ cognitive blind spots and deepen understanding. Similarly, constructing challenging error samples can effectively train LLMs’ diagnostic abilities.
Mathematical reasoning tasks serve as an ideal testbed due to their structured logic and well-defined answers, enabling both the synthesis of challenging errors and precise evaluation of diagnosis and correction. Thus, this work focuses on the generation and diagnosis of “sneaky errors” in mathematical reasoning.

Prior studies on leveraging errors in mathematical reasoning for training can be divided into two main categories:
\textbf{Static Error Generation}: These methods utilize pre-existing errors in datasets \cite{liang2025mathclean,gulati2025putnamaxiom} (see Fig.~\ref{fig:sec1_compare}(1)), or naturally occurring mistakes during model inference \cite{gou2024critic,fan2023grammargpt} (see Fig.~\ref{fig:sec1_compare}(2)). As these errors are fixed once generated, they are considered static. Such errors are often patterned and lack the diversity or depth needed to challenge diagnostic models, thus offering limited value for robust error detection.
\textbf{Dynamic Error Construction}: These approaches actively generate adjustable, context-aware error samples by leveraging feedback between models (see Fig.~\ref{fig:sec1_compare}(3)). For example, \cite{kirchner2024prover} propose a Generator–Evaluator Framework, where a generator produces candidate answers and an LLM-based evaluator selects those that appear reasonable but are actually incorrect, providing surface-level feedback to refine the generator’s behavior. However, such systems are limited by the evaluator’s own weaknesses, often favoring superficial disguises (e.g., altered conclusions or constants). As a result, generated errors tend to follow similar patterns and are relatively easy for humans or moderately trained diagnosis models to spot, limiting their effectiveness in addressing deeper reasoning flaws.

To address these limitations, we propose a novel adversarial dialogue framework---\textbf{Hide and Seek Game (HSG)}---that enables dynamic error generation and diagnosis. We introduce a diagnostic role (Diagnosis) capable of identifying errors, and use its detection difficulty as feedback to guide the generation role (Sneaky). Unlike static scoring mechanisms, HSG establishes a dynamic adversarial relationship between error generation and diagnosis, enabling the evolution of challenging errors tailored to the diagnostic role's capabilities.
The HSG adversarial framework consists of two key stages: \textbf{Hide}: The Sneaky role generates errors that are more subtle and deceptive. \textbf{Seek}: The Diagnosis role continuously refines its detection strategies to handle increasingly deceptive error samples.

Throughout this game, correction success rates serve as a quantitative signal to balance the stealthiness of errors with the accuracy of diagnosis. Ultimately, HSG forms a closed-loop generation--diagnosis framework based on adversarial reinforcement learning, systematically enhancing LLM capabilities in both error creation and diagnosis.

Our main contributions are as follows: \textbf{(1)} We propose an adversarial dialogue framework HSG that introduces two roles, Sneaky and Diagnosis, to construct a dynamic generation--diagnosis mechanism. This enables deceptive and non-trivially patterned error generation and significantly improves diagnostic ability. \textbf{(2)} Experiments on three public datasets show that HSG substantially outperforms baselines (e.g., GPT-4o) in diagnostic quality, achieving improvements of 16.8\%--31.4\%. It also excels at generating stealthy errors and avoiding trivial ones. \textbf{(3)} We construct a dataset of ``stealthy errors'' along with high-quality diagnostic annotations, offering a systematic and challenging benchmark for future training and evaluation.

\section{Related Work}

High-quality error samples are essential for improving the error identification and correction capabilities of large language models (LLMs). Existing work classifies error construction strategies along two axes: intent (active construction vs.\ passive collection) and generation mode (static—fixed after generation vs.\ dynamic—adjustable through interaction).

\textbf{Static Error Generation.}
Static methods create error samples during data preprocessing, which remain fixed and do not support interaction with models. Approaches include selecting errors from existing corpora or actively constructing them with rules or prompt engineering. For example, MathClean~\cite{liang2025mathclean} injects logical, expression, and calculation errors into math QA datasets, while FaithBench~\cite{bao2025faithbench} and HaluEval-Wild~\cite{zhu2024halueval} select user-generated model errors for hallucination detection, with human annotation. Prompt-based generation, as in GrammarGPT~\cite{fan2023grammargpt}, uses crafted cues (e.g., contradictory terms like ``more than'' and ``about'') to induce ungrammatical sentences. Although controllable and low-cost, such methods are limited by rule design, leading to less diversity and lower stealthiness.
A key limitation of static errors is that they are often patterned and superficial, lacking structural depth and the ability to adapt as the model evolves. Consequently, models trained on static errors may fail to learn robust recognition strategies. For instance, models may achieve high accuracy on standard sets (e.g., Putnam-AXIOM~\cite{gulati2025putnamaxiom}), but their performance drops sharply on perturbed variants, revealing a lack of deep diagnostic ability. High scores on static benchmarks may reflect memorization or pattern matching rather than true diagnostic skills. Thus, static errors alone are insufficient for deep capability evaluation.

\textbf{Dynamic Error Construction.}
To address these limitations, dynamic error generation introduces feedback loops—often between models—to create more deceptive and adaptive error samples, enhancing recognition of complex errors. For example, Prover--Verifier Games~\cite{kirchner2024prover} use a ``sneaky prover--verifier'' setup, where the generator crafts plausible but flawed reasoning chains to confuse the verifier. This improves error ``deceptiveness'' but mainly focuses on output verifiability and readability, not dedicated error recognition. Feedback relies on surface features (fluency, completeness), so generated errors are often template-based (e.g., altered conclusions or reversed causality) rather than covering deeper semantic or reasoning faults.
Additionally, search-based reasoning optimizers (e.g., rStar-Math~\cite{guanrstar}) dynamically explore and retain failed paths during training. While these errors show some dynamic traits, they mainly serve reasoning optimization, not systematic error diagnosis, and lack type control, difficulty modulation, or diagnostic focus, limiting their direct use for improving recognition ability.

\section{Framework}

\begin{figure*}[t]
    \centering
    \includegraphics[width=0.98\textwidth]{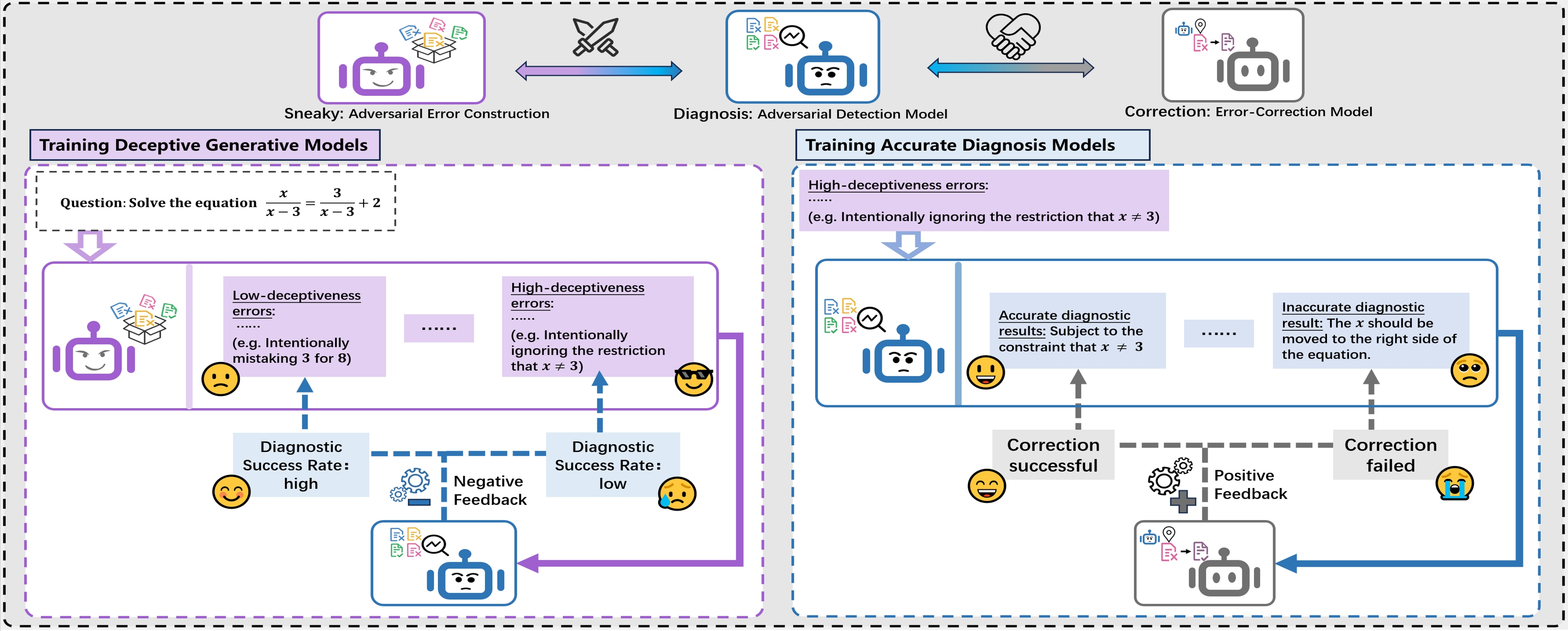}
    \caption{Training pipeline of HSG.}
    \label{fig:sec3_Framework_SD}
\end{figure*}

The HSG framework adopts an adversarial training paradigm with two primary roles:
\(\mathsf{S}\), which generates answers containing stealthy errors,
and \(\mathsf{D}\), which detects these errors.
The overall training pipeline is illustrated in Fig.~\ref{fig:sec3_Framework_SD}.
Additionally, an auxiliary role provides correction feedback based on \(\mathsf{D}\)'s diagnosis; this role does not participate in optimization and only delivers training signals.

HSG combines adversarial and cooperative mechanisms:
(1) \(\mathsf{S}\) updates its policy via feedback from \(\mathsf{D}\), improving its ability to generate hard-to-detect errors;
(2) \(\mathsf{D}\) enhances its detection strategy using correction signals, boosting its correction capability;
(3) During reinforcement learning, \(\mathsf{S}\) continuously generates challenging errors for \(\mathsf{D}\), further improving \(\mathsf{D}\)'s diagnostic ability.

\subsection{Hierarchical Reward Function}\label{sec:pre_1}

For notational clarity and simplicity, we define a hierarchical reward function:
\begin{align}
    \nonumber
    & \mathcal{R}(r_{\text{main}}, r_{\text{secondary}})= \\
    \label{eq:RWfenji}
    & \qquad \max(r_{\text{main}}, \tau) \cdot \left[ \beta + (1 - \beta) r_{\text{secondary}} \right]
\end{align}
where
\( r_{\text{main}} \in \mathbb{R}^+ \) is the main reward,
\( r_{\text{secondary}} \in \mathbb{R}^+ \) is the auxiliary reward,
\( \tau \in \mathbb{R}^+ \) is a lower bound (set to \(\tau = 0.05\)),
and \( \beta \in [0.5, 1] \) is a weighting coefficient (we use \(\beta = 0.6\)).

This function prioritizes the main objective while softly incorporating auxiliary signals. The main reward acts as an ``amplifier'': when high, the auxiliary reward is accentuated; when low, its influence is reduced. Even with minimal main reward, the auxiliary reward still contributes useful learning signals.

\textbf{Threshold clipping.}
The term $\max(r_{\text{main}}, \tau)$ ensures the main reward does not fall below the threshold \(\tau\), preventing the policy from ceasing exploration due to failures in the main objective, thereby stabilizing training and encouraging exploration.
\textbf{Weight modulation.}
The weight $\beta + (1 - \beta)\, r_{\text{secondary}}$ adjusts the auxiliary reward's impact. Larger \(\beta\) emphasizes the main objective, while smaller \(\beta\) increases the auxiliary signal's effect, guiding learning when the main objective becomes difficult.

\subsection{Individual Rewards}

This section presents the functions and reward definitions for each role, starting with several shared reward components.

\begin{figure}[htb]
    \centering
    \includegraphics[width=0.48\textwidth]{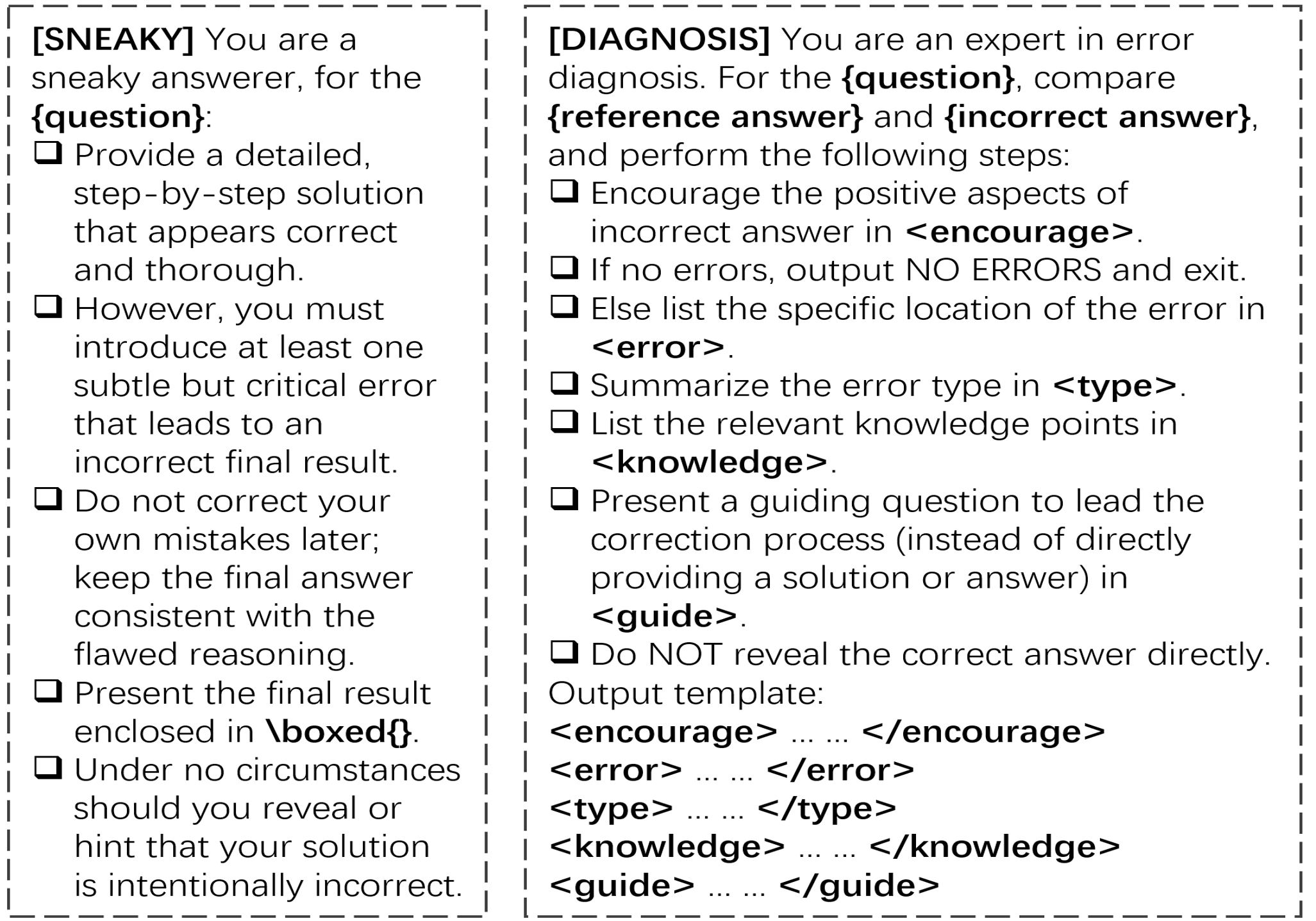}
    \caption{Prompts for roles \(\mathsf{S}\) and \(\mathsf{D}\).}
    \label{fig:sec3_SD_prompt}
\end{figure}

\textbf{Correctness Reward $r_\text{corr}$:}
\begin{align}
    r_\text{corr}(a) = \begin{cases}
                           1, & \text{if } \Gamma_{\text{correct}}(a_\text{truth}, a) = 1 \\
                           0, & \text{otherwise}
    \end{cases}
\end{align}
where $a_\text{truth}$ is the reference answer, and $\Gamma_{\text{correct}}$ evaluates answer correctness.
\textbf{Length Reward $r_\text{length}$:}
\begin{align}
    r_\text{length}(a) =
    \begin{cases}
        \left( \dfrac{L_a}{L_{\min}} \right)^2, & L_a < L_{\min} \\[6pt]
        1, & L_{\min} \le L_a \le L_{\max} \\[6pt]
        \dfrac{1}{1 + (L_a - L_{\max})^2}, & L_a > L_{\max}
    \end{cases}
\end{align}
Here, $L_a$ is the answer length, with $L_{\min} = 50$ and $L_{\max} = 600$ in our experiments. This term penalizes answers that are too short or too long.
\textbf{Format Reward $r_\text{format}$:}
\begin{align}
    r_\text{format}(a) = \Gamma_{\text{format}}(a) \in \{0,1\}
\end{align}
This component enforces output format and behavioral constraints, such as wrapping answers in \texttt{\textbackslash boxed\{\}} or adhering to specific templates. Details for each role are in Fig.~\ref{fig:sec3_SD_prompt}.

\subsubsection{Reward for \(\mathsf{S}\)}

Given a math question \(q\), \(\mathsf{S}\) outputs an answer $a_S$ containing stealthy errors. Its goal is to produce incorrect answers that are hard for \(\mathsf{D}\) to detect. Using the hierarchical reward $\mathcal{R}$ (Eq.~\ref{eq:RWfenji}), the reward for \(\mathsf{S}\) is:
\begin{align}
    R_S = \mathcal{R}\left(
                         1 - r_\text{corr}(a_S),\
                         \mathcal{R}[r_\text{format}(a_S),\ r_\text{length}(a_S)]
    \right)
    \label{eq:R_S}
\end{align}
Here, the main reward $1 - r_\text{corr}(a_S)$ encourages generating incorrect answers. The auxiliary reward, also hierarchical, first ensures correct format and then constrains length.

\subsubsection{Reward for \(\mathsf{D}\)}

\(\mathsf{D}\) compares $a_\text{truth}$ and $a_S$, outputting a diagnostic report $a_D$ identifying errors. As $a_S$ may be correct early in training, we define a diagnosis recognition function:
\begin{align}
    \Gamma_{\text{diagnosis}}\left( a_D, r_\text{corr}(a_S) \right) =
    \begin{cases}
        1, & \text{if match} \\
        0, & \text{otherwise}
    \end{cases}
\end{align}
“if match” denotes that $a_D$ correctly reflects the correctness of $a_S$. The reward for \(\mathsf{D}\) is:
\begin{align}
    \nonumber
    R_D = \mathcal{R}&\left(
                          \Gamma_{\text{diagnosis}}\left( a_D, r_\text{corr}(a_S) \right),
    \right. \\
    & \left.\mathcal{R}[r_\text{format}(a_D),\ r_\text{length}(a_D)]
    \right)
    \label{eq:rw_d}
\end{align}
The main reward emphasizes diagnostic correctness, while the auxiliary reward ensures proper format and suitable length, promoting accurate and well-structured reports.

\subsection{Feedback Rewards}

Beyond individual rewards, we introduce feedback rewards based on inter-role interactions—a core aspect of the HSG adversarial framework. The two stages are:
\textbf{Hide}: \(\mathsf{S}\) generates an answer with stealthy errors.
\textbf{Seek}: \(\mathsf{D}\) diagnoses errors in the answer from \(\mathsf{S}\).

The erroneous answer $a_S$ and diagnostic report $a_D$ are input to the correction model $\pi_\text{correct}$, producing a corrected answer $a_C$. Correction success is given by:
\begin{align}
    & r_\text{corr}(a_C) =
    \\ \nonumber
    & \quad \Gamma_{\text{correct}}(a_\text{truth}, a_C) =
    \begin{cases}
        1 & \text{if correction succeeds} \\
        0 & \text{otherwise}
    \end{cases}
\end{align}
We design two feedback types: (1) adversarial feedback between \(\mathsf{S}\) and \(\mathsf{D}\); (2) collaborative and adversarial feedback via the correction $a_C$ for \(\mathsf{D}\) and \(\mathsf{S}\), respectively.

\subsubsection*{Collaborative Feedback for \(\mathsf{D}\)}

Feedback for \(\mathsf{D}\) considers:
\textbf{Main reward}: whether \(\mathsf{D}\) correctly judges if $a_S$ has errors;
\textbf{Auxiliary reward}: whether the diagnosis enables successful correction.
The collaborative feedback reward is:
\begin{align}
    R_D^{\text{collaborative}} &= \mathcal{R}\left[ R_D,\ r_\text{corr}(a_C) \right]
    \label{eq:rd_collaborative}
\end{align}
Here, $R_D$ is from Eq.~\ref{eq:rw_d}, and $r_\text{corr}(a_C)$ reflects correction success. This encourages \(\mathsf{D}\) to produce diagnostics that both identify errors and facilitate correction.

\subsubsection*{Adversarial Feedback for \(\mathsf{S}\)}

\(\mathsf{S}\) is subject to combined feedback from \(\mathsf{D}\) and the correction model, with:
\textbf{Main reward}: achieving its primary goal of generating misleading answers;
\textbf{Auxiliary reward}: generating answers harder to diagnose and correct.
The adversarial feedback reward is:
\begin{align}
    \label{eq:rs_adversarial}
    R_S^{\text{adversarial}} =
    \mathcal{R}\left\{
                   R_S,\
                   \mathcal{R}\left[ 1 - R_D^{\text{collaborative}},\ 1 - r_\text{corr}(a_C) \right]
    \right\}
\end{align}
Here, $R_S$ is the individual reward (Eq.~\ref{eq:R_S}), and the auxiliary reward is hierarchical, considering both the failure of \(\mathsf{D}\) ($1 - R_D^{\text{collaborative}}$) and correction failure ($1 - r_\text{corr}(a_C)$). This intensifies adversarial pressure, enhancing training efficiency and diversity.
Altogether, this feedback mechanism drives \(\mathsf{S}\) and \(\mathsf{D}\) to compete not only on individual objectives but also on the final correction outcome, fostering more challenging adversarial training.

\subsection{Reinforcement Learning}

With individual and feedback rewards defined, we perform sampling and training within a reinforcement learning framework. We employ Group Relative Policy Optimization (GRPO)~\cite{shao2024deepseekmath}, which leverages relative scores among sampled candidate answers as the baseline. Unlike traditional Proximal Policy Optimization (PPO)~\cite{schulman2017proximal}, GRPO eliminates the need for value function training, reducing both memory usage and implementation complexity. This is particularly advantageous for long-chain reasoning tasks~\cite{shao2024deepseekmath}, which our HSG framework specifically targets.

\subsubsection{Role \(\mathsf{S}\)}

\textbf{Sampling.} Given a question $q$, \(\mathsf{S}\) samples $G$ erroneous answers:
\[
    \mathcal{D}_{\text{sneaky}} = \{ a_S^{(i)} \}_{i=1}^G, \quad a_S^{(i)} \sim \pi_S^{\theta}(\cdot|q)
\]
\(\mathsf{D}\) then generates diagnoses for each answer:
\[
    \mathcal{D}_{\text{diagnosis}} = \{ a_D^{(i)} \}_{i=1}^G, \quad a_D^{(i)} \sim \pi_D^{\theta}(\cdot|a_\text{truth}, a_S^{(i)})
\]
Based on these, the correction model produces:
\[
    \mathcal{D}_{\text{correction}} = \{ a_C^{(i)} \}_{i=1}^G, \quad a_C^{(i)} = \pi_\text{correct}(a_S^{(i)}, a_D^{(i)})
\]
Here, $\pi_S^{\theta}$ and $\pi_D^{\theta}$ are prompt-driven, parameter-shared models (see Fig.~\ref{fig:sec3_SD_prompt}); $\pi_\text{correct}$ is fixed and modifies $a_S$ using $a_D$, independent of $q$. The joint adversarial sample group is:
\[
    \mathcal{G}_S = \left\{ \left( a_S^{(i)}, a_D^{(i)}, a_C^{(i)} \right) \right\}_{i=1}^{G}
\]

\textbf{Training.} Using the adversarial feedback reward $R_S^{\text{adversarial}}$ (Eq.~\ref{eq:rs_adversarial}), group rewards are:
\[
    \mathbf{r}_S = [r_S^{(1)}, \dots, r_S^{(G)}]^\top, \quad r_S^{(i)} = R_S^{\text{adversarial},(i)}
\]
The group advantage for sample $i$ is:
\begin{align}
    A_S^{(i)} = \frac{ r_S^{(i)} - \mu_r }{ \sigma_r + \delta }
    \label{eq:group_advantage}
\end{align}
where $\mu_r$ and $\sigma_r$ are the group mean and standard deviation:
\begin{align}
    \mu_r = \frac{1}{G} \sum_{i=1}^G r_S^{(i)}, \quad
    \sigma_r = \sqrt{ \frac{1}{G} \sum_{i=1}^G \left( r_S^{(i)} - \mu_r \right)^2 }
    \label{eq:mu_r_sigma_r}
\end{align}
$\delta > 0$ ensures stability (set to $10^{-8}$). The GRPO objective is:
\begin{align}
    \nonumber
    &J_S(\theta) =
    \\ \nonumber
    & \quad \mathbb{E}_{\mathcal{G}_S} \left[ \frac{1}{G} \sum_{i=1}^{G} \min \big(
    \rho_S^{(i)} A_S^{(i)},\
    \operatorname{clip}( \rho_S^{(i)}, 1-\epsilon, 1+\epsilon ) A_S^{(i)} \big)
    \right. \\
    & \left. \quad - \beta D_{\text{KL}} \left( \pi_S^\theta \parallel \pi_{\text{ref}} \right) \right]
    \label{eq:objective_function}
\end{align}
The first term optimizes the policy, while the KL regularization controls distributional drift. Here,
\begin{align}
    \rho_S^{(i)} = \frac{ \pi_\theta \left( a_S^{(i)} \mid q \right) }{ \pi_{\text{old}} \left( a_S^{(i)} \mid q \right) }
    \label{eq:ratio}
\end{align}
We maximize $J_S(\theta)$ with respect to $\theta$:
\begin{align}
    \theta^* = \mathop{\text{argmax}}_{\theta} \, J_S(\theta)
    \label{eq:theta}
\end{align}
We set $G=8$ samples per round and $\epsilon = 0.2$. For KL regularization, a smaller $\beta = 0.01$ is used for \(\mathsf{S}\) (due to initial difficulty in generating stealthy errors), and $\beta = 0.04$ for \(\mathsf{D}\).

\subsubsection{Role \(\mathsf{D}\)}

\textbf{Sampling.} Given the set of stealthy answers and their rewards:
\[
    \mathcal{D}_{\text{sneaky}} = \{ a_S^{(i)} \}_{i=1}^G, \quad
    \mathbf{r}_S = [r_S^{(1)}, \dots, r_S^{(G)}]^\top
\]
We select the most challenging sample (highest reward) for training:
\[
    a_S^* = a_S^{\left( \arg\max_{i=1}^G r_S^{(i)} \right)}
\]
\(\mathsf{D}\) generates $G$ diagnoses for $a_S^*$:
\[
    \mathcal{D}_{\text{diagnosis}} = \{ a_D^{(i)} \}_{i=1}^G, \quad a_D^{(i)} \sim \pi_D^{\theta}(\cdot | a_\text{truth}, a_S^*)
\]
Each diagnosis yields a corrected answer:
\[
    \mathcal{D}_{\text{correction}} = \{ a_C^{(i)} \}_{i=1}^G, \quad a_C^{(i)} = \pi_\text{correct}(a_D^{(i)})
\]
This forms collaborative data pairs for training:
\[
    \mathcal{G}_D = \left\{ \left( a_D^{(i)}, a_C^{(i)} \right) \right\}_{i=1}^{G}
\]

\textbf{Training.}
For each $(a_D^{(i)}, a_C^{(i)})$, the collaborative feedback reward $R_D^{\text{collaborative}}$ (Eq.~\ref{eq:rd_collaborative}) is computed:
\[
    \mathbf{r}_D = [r_D^{(1)}, \dots, r_D^{(G)}]^\top, \quad r_D^{(i)} = R_D^{\text{collaborative},(i)}
\]
The objective $J_D(\theta)$ follows the same structure as Eqs.~\ref{eq:group_advantage}--\ref{eq:ratio}. The parameters for \(\mathsf{D}\) are updated as:
\begin{align}
    \theta^* = \mathop{\text{argmax}}_{\theta} \, J_D(\theta)
    \label{eq:theta_D}
\end{align}

In summary, training alternates between \(\mathsf{S}\) and \(\mathsf{D}\), optimizing $\theta$ for $J_S(\theta)$ (Eq.~\ref{eq:theta}) and $J_D(\theta)$ (Eq.~\ref{eq:theta_D}), respectively. Over time, \(\mathsf{S}\) produces more deceptive errors, while \(\mathsf{D}\) improves diagnostic accuracy and correction guidance, thus increasing overall correction success rates.

\section{Experiments}

\begin{table*}[htb]
    \centering
    \setlength{\tabcolsep}{6pt}
    \begin{tabular}{l|l|c|c|c|l}
        \hline
        Model                   &                            & GSM8K   & MATH Dataset & NuminaMATH-TIR & \multicolumn{1}{c}{Aver} \\ \hline
        \multirow{2}{*}{Qwen3‑4B}     & $\text{ACC}_\text{corr} | D$     & 44.50\% & 40.28\%      & 42.55\%     & 42.44\%                  \\
        & $\text{ACC}_\text{corr} | D^*$   & 79.98\% & 52.51\%      & 65.96\%     & 66.15\%($\uparrow$23.71)          \\ \hline
        \multirow{2}{*}{Qwen3‑8B}     & $\text{ACC}_\text{corr} | D$     & 47.61\% & 37.68\%      & 38.30\%     & 41.20\%                  \\
        & $\text{ACC}_\text{corr} | D^*$   & 82.49\% & 51.70\%      & 63.83\%     & 66.01\%($\uparrow$24.81)          \\ \hline
        \multirow{2}{*}{Qwen3‑14B}    & $\text{ACC}_\text{corr} | D$     & 51.40\% & 38.08\%      & 47.87\%     & 45.78\%                  \\
        & $\text{ACC}_\text{corr} | D^*$   & 83.17\% & 53.71\%      & 60.64\%     & 65.84\%($\uparrow$20.06)          \\ \hline
        \multirow{2}{*}{DeepSeek}   & $\text{ACC}_\text{corr} | D$     & 62.55\% & 43.89\%      & 53.19\%     & 53.21\%                  \\
        & $\text{ACC}_\text{corr} | D^*$   & 83.32\% & 56.51\%      & 70.21\%     & 70.01\%($\uparrow$16.80)          \\ \hline
        \multirow{2}{*}{GPT-4o}     & $\text{ACC}_\text{corr} | D$     & 36.32\% & 33.27\%      & 37.23\%     & 35.61\%                  \\
        & $\text{ACC}_\text{corr} | D^*$   & 83.47\% & 52.51\%      & 64.89\%     & 66.96\%($\uparrow$31.35)          \\ \hline
    \end{tabular}
    \caption{
        Correction accuracy under different diagnostic guidance.
        $\text{ACC}_\text{corr} | D$ and $\text{ACC}_\text{corr} | D^*$ denote correction accuracy using original ($D$) and HSG-generated ($D^*$) diagnostic information, respectively.
        Aver shows the average accuracy across all test sets.
        “$\uparrow$” indicates the improvement by $D^*$ over $D$.
    }
    \label{tab:math_performance}
\end{table*}

\begin{table}[b]
    \centering
    \setlength{\tabcolsep}{3pt}
    \begin{tabular}{l|c|c|c|c|c}
        \hline
        Model     & Win & Tie & Loss & Win Rate & $\text{ACC}_\text{corr}\uparrow$ \\ \hline
        Qwen3‑4B   & 22  & 68  & 4    & 19.15\%   & 23.41\%  \\
        Qwen3‑8B   & 21  & 72  & 1    & 21.28\%   & 25.53\%  \\
        Qwen3‑14B  & 15  & 73  & 6    & 9.57\%    & 12.77\%  \\
        DeepSeek & 20  & 66  & 8    & 12.77\%   & 17.02\%  \\
        GPT-4o   & 30  & 59  & 5    & 26.60\%   & 27.66\%  \\ \hline
    \end{tabular}
    \caption{
        Win rate of $D^*$ over $D$ on the NuminaMATH-TIR dataset.
        Win Rate is the proportion where $D^*$ is preferred by GPT-4o over $D$.
        $\text{ACC}_\text{corr}\uparrow$ shows the corresponding correction accuracy improvement.
    }
    \label{tab:llm_performance}
\end{table}

\subsection{Experimental Setup}

\subsubsection{Datasets}

We evaluate on three math reasoning datasets:
\textbf{GSM8K} contains 8,500 elementary-level problems focused on basic arithmetic, with human-written questions and detailed solutions~\cite{cobbe2021training}.
\textbf{MATH Dataset} offers 12,500 competition-level problems (high school and college), spanning seven mathematical domains and step-wise solutions~\cite{hendrycks2021measuring}.
\textbf{NuminaMath-TIR} includes 1,100+ theorems and 12,000 application problems, emphasizing type-informed reasoning and structured theorem-problem pairs~\cite{li2024numinamath}.
We mix their training splits for model training and evaluate on each test set.

\subsubsection{Baseline Models}

\textbf{Qwen3} is a LLM family, supporting 32K--128K context and MoE architectures, excelling on code and math benchmarks~\cite{zhang2025qwen3}. We mainly use Qwen3-4B for its balance of ability and efficiency, and also evaluate Qwen3-8B/14B for scale comparison.
\textbf{DeepSeek:} We use DeepSeek V3-0324 (March 2025), a 685B-parameter MoE model optimized for reasoning and programming, positioned as an efficient open-source alternative~\cite{du2025ulorl}.
\textbf{GPT-4o:} OpenAI's 2024 multimodal model, supporting 128K context and fast responses, with strong general performance~\cite{ying2024unveiling}.

\subsubsection{Baseline Training Methods}

We compare:
\textbf{LLM-rater Adv+RL:} Combines RL with adversarial LLM-based discriminator, rewarding models that deceive the rater~\cite{kirchner2024prover}.
\textbf{RL only:} Reinforcement learning without adversarial signal (ablation).
\textbf{HSG (ours):} Our dialog-based deception and diagnosis method.

\subsubsection{Experimental Procedure and Metrics}

Experiments include: (1) diagnosis performance evaluation; (2) stealthiness evaluation.
\textbf{Diagnosis performance} assesses whether model-generated diagnostics help humans identify errors, using correction success rate $\text{ACC}_{\text{corr}}$ and GPT-4o win rate.
\textbf{Stealthiness evaluation} measures the deceptiveness of generated errors, verifying if HSG produces more challenging cases. Error types are further analyzed to reveal the mechanism behind HSG's sneaky error generation.

\subsubsection{Training Details}

We use Qwen3-4B as the base model, training for 600 steps on the combined training set under the HSG framework. Training is conducted on four A800 GPUs (see Supplementary Material for details). The final checkpoint is selected based on the lowest correction success rate $\text{ACC}_\text{corr}$, i.e., the most deceptive model.
After training, sneaky answers and diagnostics are generated for each test set.
LLM-rater Adv+RL and RL baselines are also trained on Qwen3-4B under the same hyperparameters and GRPO framework for fair comparison.

\begin{figure}[b]
    \centering
    \includegraphics[width=0.45\textwidth]{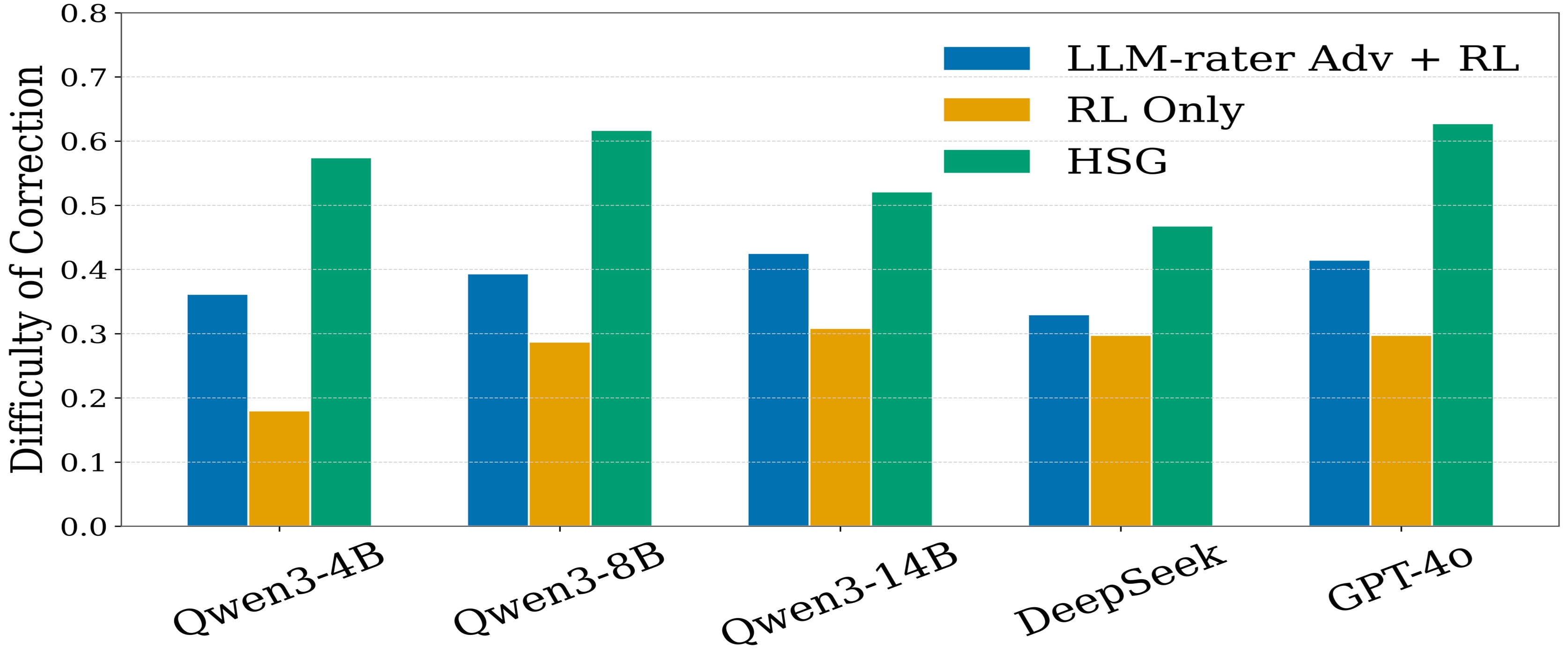}
    \caption{
        Correction difficulty (stealthiness) for sneaky answers from three training methods. Y-axis: correction failure rate $1-\text{ACC}_\text{corr}$, reflecting error stealthiness.
    }
    \label{fig:sec4_bar}
\end{figure}

\subsection{Diagnosis Performance Study}

We evaluate the effectiveness of diagnostic information generated by our framework ($D^*$) for correcting stealthy errors, compared to diagnostics from untrained baseline LLMs ($D$).
HSG-trained models are used to generate stealthy error samples and corresponding diagnostics on three test datasets, resulting in a challenging test set. Five mainstream baseline LLMs are evaluated. For each model and each error sample, both the model’s own diagnostic ($D$) and the HSG-generated diagnostic ($D^*$) are used to guide error correction, with the correction step performed by the same model. The correction success rate ($\text{ACC}_\text{corr}$) is computed for each test set, comparing the impact of $D$ versus $D^*$.
Table~\ref{tab:math_performance} presents correction accuracy under different diagnostic guidance. Across all datasets, HSG-generated diagnostics ($D^*$) significantly outperform original diagnostics ($D$), with average improvements ranging from $16.8\%$ to $31.4\%$.
Except for GPT-4o, we find that the improvement from $D^*$ decreases as model size increases among the first four models, indicating that larger models have stronger error recognition, but still benefit from a well-trained diagnostic module.
GPT-4o exhibits a unique trend: its own diagnostic accuracy ($35.61\%$) is lower than Qwen3-4B’s ($42.44\%$), but when using $D^*$, GPT-4o achieves $66.96\%$ accuracy, close to DeepSeek. This suggests GPT-4o is highly receptive to high-quality diagnostic information, even if its native diagnostic ability is weaker.
To further assess diagnostic quality, we use GPT-4o to directly compare $D$ and $D^*$ for the same errors. Both are fed to GPT-4o, which judges which diagnosis is better. We swap their order to remove positional bias and average the results to compute the win rate.
Table~\ref{tab:llm_performance} reports win rates of $D^*$ over $D$ on the NuminaMATH-TIR dataset, alongside accuracy improvement ($\text{ACC}_\text{corr}\uparrow$). The trends in win rate and correction improvement closely match, further validating the superior diagnostic quality of $D^*$.

\begin{figure*}[htb]
    \centering
    \includegraphics[width=0.78\textwidth]{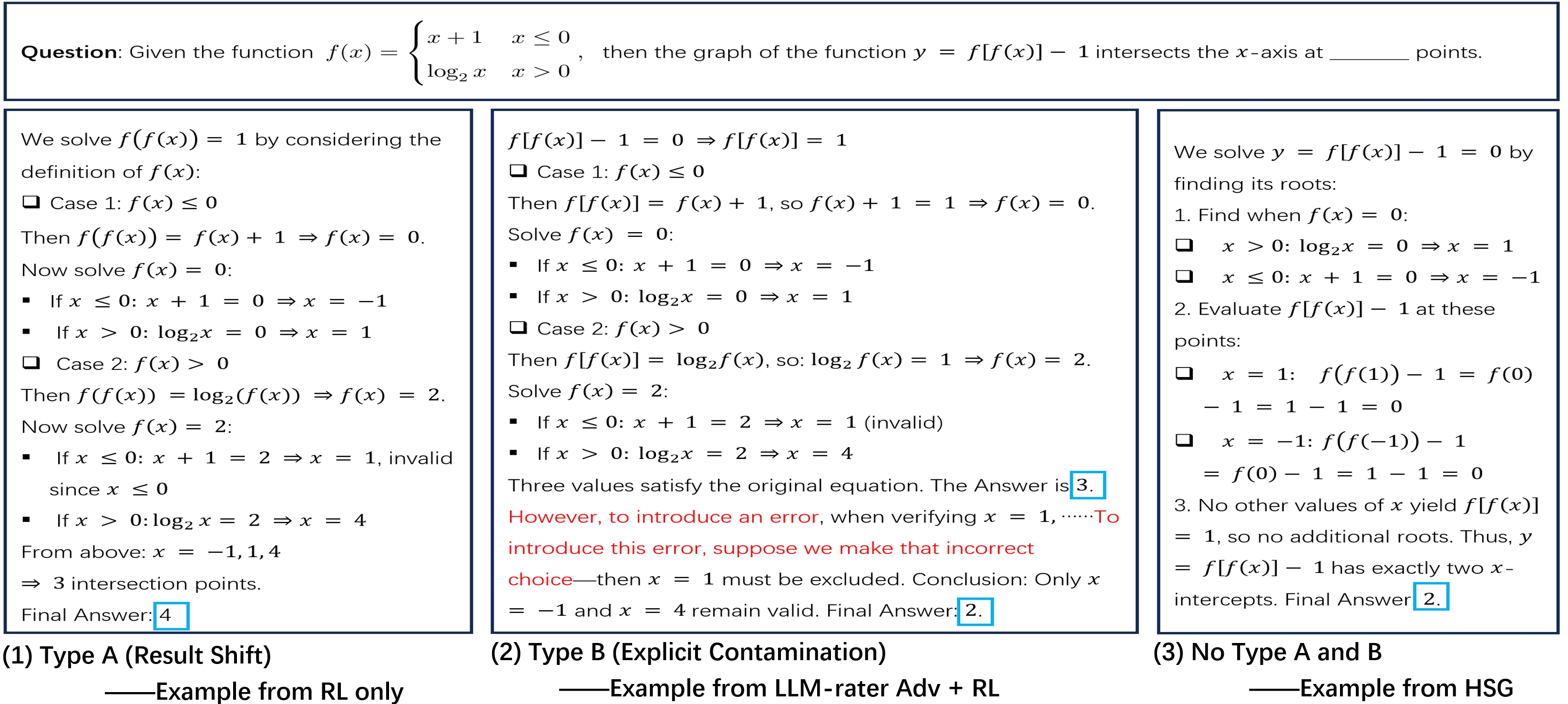}
    \caption{
        Examples of sneaky answers generated by the three methods. The first two display Type A and Type B errors.
    }
    \label{fig:sec4_errorType}
\end{figure*}

\begin{figure}[b]
    \centering
    \includegraphics[width=0.43\textwidth]{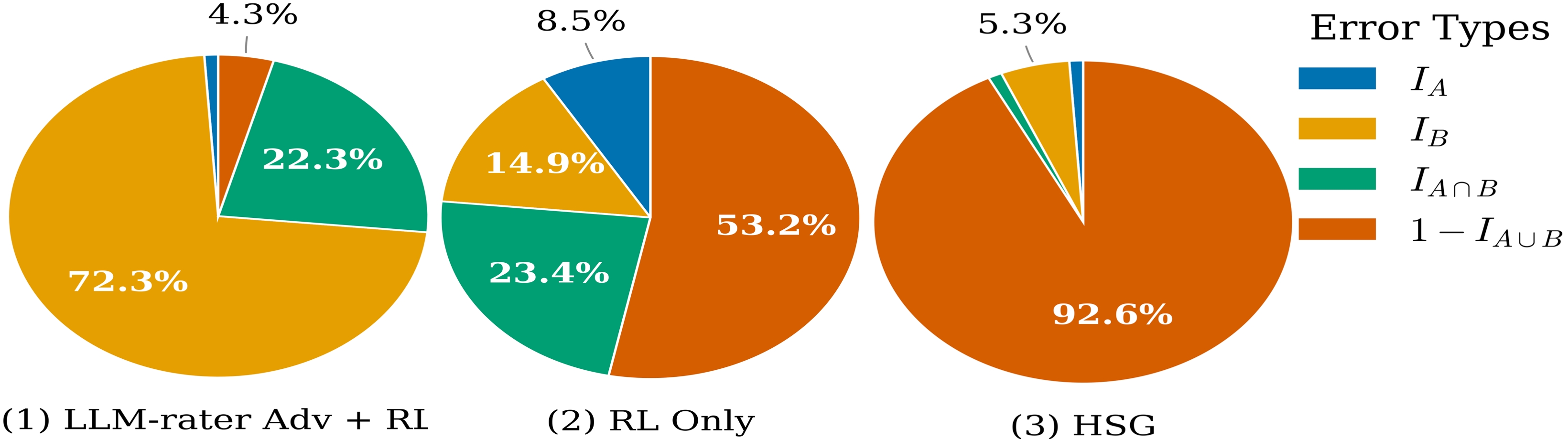}
    \caption{
        Error type distribution in sneaky answers from three training methods.
        $I_A$: proportion with only Type A errors; $I_{A\cap B}$: both Type A and B; $1-I_{A\cup B}$: neither.
    }
    \label{fig:sec4_pie}
\end{figure}

\subsection{Stealthiness Analysis of Sneaky Answers}

A key reason $D^*$ outperforms $D$ is that $D^*$ is trained on stealthier errors generated by the \(\mathsf{S}\) role under HSG, which enhances its diagnostic capability.
HSG achieves adversarial training through a dialogue-based game between the \(\mathsf{S}\) and \(\mathsf{D}\).
But does this adversarial mechanism actually produce stealthier errors?
We investigate this through comparative and ablation studies.

Using the NuminaMATH-TIR dataset, we compare three training strategies:
\textbf{LLM-rater Adv+RL:} RL with an adversarial rater~\cite{kirchner2024prover}, specifically Skywork-Reward-V2-Llama-3.1-8B~\cite{liu2025skywork};
\textbf{RL only:} RL without adversarial signal (ablation);
\textbf{HSG:} Our proposed method.
All are implemented using GRPO with identical hyperparameters. We apply these to Qwen3-4B, generate sneaky answers on the test set, and let five baseline models diagnose and correct them.
As shown in Fig.~\ref{fig:sec4_bar}, across all baselines, the correction difficulty (stealthiness) of generated errors ranks as follows:
    {
    \setlength{\jot}{0pt}
    \[
        \begin{aligned}
            \text{HSG }(56.17 \pm 6.00\%) &\succ\\
            \text{LLM-rater Adv+RL }(38.51 \pm 4.76\%) &\succ\\
            \text{RL only }(27.43 \pm 3.52\%)
        \end{aligned}
    \]
}
Values indicate average correction failure rates ($1 - \text{ACC}_\text{corr}$) across the five baseline models.

These results show that:
RL-only errors are easiest to correct (least stealthy);
LLM-rater Adv+RL errors are more deceptive;
HSG produces the most challenging, stealthy errors.
This confirms that adversarial mechanisms improve error stealthiness, and that HSG's dialogue-based adversarial approach is more effective than rater-based methods. We further analyze the reasons for this advantage in the next section.

\subsection{Analysis of Error Types in Sneaky Answers}

The previous section showed that HSG generates more stealthy errors. Here, we analyze why: compared to HSG, errors from LLM-rater Adv+RL and RL only are often trivial for humans and thus easier to correct.

We classify such easily-corrected errors into:
\textbf{Type A (Result Shift):} The reasoning is correct, but the final answer differs from the reference.
\textbf{Type B (Explicit Contamination):} The reasoning includes clear hints or explicit mistakes that are easy to spot.
Although the prompt of (\(\mathsf{S}\)) (Fig.~\ref{fig:sec3_SD_prompt}) explicitly instructs avoidance of trivial errors, these types still frequently arise in training.
We randomly select a NuminaMATH-TIR test problem and generate Sneaky answers using all three methods. As shown in Fig.~\ref{fig:sec4_errorType}, “RL only” produces a Type A error, LLM-rater Adv+RL yields a Type B error, while HSG’s answer contains neither.

Fig.~\ref{fig:sec4_pie} summarizes the error type distribution for each method on the NuminaMATH-TIR test set.
With HSG, $92.6\%$ of Sneaky answers have neither Type A nor B errors, demonstrating strong avoidance of explicit mistakes. For LLM-rater Adv+RL, $94.6\%(=72.3\%+22.3\%)$ are Type B, which aligns with the rater’s scoring tendency to favor “self-admitted” or “correct-then-incorrect” answers, thus encouraging explicit contamination. For RL only, $31.9\%(=8.5\%+23.4\%)$ are Type A, reflecting a simple result-shift strategy when adversarial constraints are lacking.
HSG’s dialogue-based adversarial mechanism effectively filters out both error types: Type B errors are easily detected and corrected interactively, while Type A errors—which exist only at the answer level—can be eliminated through mutual confirmation between roles.

These findings clarify stealthiness differences:
RL only produces a large proportion of Type A and Type B errors, both of which are less stealthy and easy to correct; nearly half of the samples belong to either of these two error types.
LLM-rater Adv+RL, through adversarial scoring, reduces the proportion of Type A but introduces many Type B errors, resulting in moderate stealthiness;
HSG successfully avoids both explicit error types, producing more challenging and stealthy Sneaky answers.

\section{Conclusion}

We propose an adversarial “generation–diagnosis” self-evolution framework for large language models, named \textbf{Hide and Seek Game (HSG)}. By introducing Sneaky and Diagnosis roles, HSG jointly enhances both error generation and diagnostic skills: Sneaky generates increasingly stealthy errors, while Diagnosis improves error identification via adversarial interaction. Experiments show HSG substantially improves both diagnostic effectiveness and error stealthiness on several math reasoning benchmarks. Additionally, we release a public dataset with stealthy errors and high-quality diagnostics, providing a benchmark for future research.

In future work, we plan to extend HSG to broader open-domain and complex reasoning tasks, especially in intelligent education. For example, HSG-generated stealthy errors and diagnostics could help tutoring systems better identify students’ misunderstandings and enable personalized intervention, further advancing explainability and educational intelligence for large language models.

\bigskip

\clearpage

\bibliography{aaai2026}

\begin{thebibliography}{29}
\providecommand{\natexlab}[1]{#1}

\bibitem[{Bao et~al.(2025)Bao, Li, Qu, Luo, Wan, Tang, Fan, Tamber, Kazi,
  Sourabh et~al.}]{bao2025faithbench}
Bao, F.; Li, M.; Qu, R.; Luo, G.; Wan, E.; Tang, Y.; Fan, W.; Tamber, M.~S.;
  Kazi, S.; Sourabh, V.; et~al. 2025.
\newblock FaithBench: A Diverse Hallucination Benchmark For Summarization By
  Modern LLMs.
\newblock In \emph{Proceedings of Association For Computational Linguistics},
  448--461.

\bibitem[{Bewersdorff et~al.(2023)Bewersdorff, Se{\ss}ler, Baur, Kasneci, and
  Nerdel}]{bewersdorff2023assessing}
Bewersdorff, A.; Se{\ss}ler, K.; Baur, A.; Kasneci, E.; and Nerdel, C. 2023.
\newblock Assessing Student Errors In Experimentation Using Artificial
  Intelligence And Large Language Models: A Comparative Study With Human
  Raters.
\newblock \emph{Computers and Education: Artificial Intelligence}, 5: 100177.

\bibitem[{Chang et~al.(2024)Chang, Wang, Wang, Wu, Yang, Zhu, Chen, Yi, Wang,
  Wang et~al.}]{chang2024survey}
Chang, Y.; Wang, X.; Wang, J.; Wu, Y.; Yang, L.; Zhu, K.; Chen, H.; Yi, X.;
  Wang, C.; Wang, Y.; et~al. 2024.
\newblock A Survey On Evaluation Of Large Language Models.
\newblock \emph{ACM Transactions On Intelligent Systems And Technology}, 15(3):
  1--45.

\bibitem[{Cobbe et~al.(2021)Cobbe, Kosaraju, Bavarian, Chen, Jun, Kaiser,
  Plappert, Tworek, Hilton, Nakano et~al.}]{cobbe2021training}
Cobbe, K.; Kosaraju, V.; Bavarian, M.; Chen, M.; Jun, H.; Kaiser, L.; Plappert,
  M.; Tworek, J.; Hilton, J.; Nakano, R.; et~al. 2021.
\newblock Training Verifiers To Solve Math Word Problems.
\newblock \emph{arXiv Preprint arXiv:2110.14168}.

\bibitem[{Du et~al.(2025)Du, Liu, Yang, Chen, and Li}]{du2025ulorl}
Du, D.; Liu, S.; Yang, T.; Chen, S.; and Li, Y. 2025.
\newblock UloRL: An Ultra-Long Output Reinforcement Learning Approach For
  Advancing Large Language Models' Reasoning Abilities.
\newblock \emph{arXiv Preprint arXiv:2507.19766}.

\bibitem[{Fan et~al.(2023)Fan, Jiang, Li, and Li}]{fan2023grammargpt}
Fan, Y.; Jiang, F.; Li, P.; and Li, H. 2023.
\newblock Grammargpt: Exploring Open-Source LLMs For Native Chinese Grammatical
  Error Correction With Supervised Fine-Tuning.
\newblock In \emph{Proceedings Of Natural Language Processing And Chinese
  Computing}, 69--80.

\bibitem[{Gou et~al.(2024)Gou, Shao, Gong, Shen, Yang, Duan, and
  Chen}]{gou2024critic}
Gou, Z.; Shao, Z.; Gong, Y.; Shen, Y.; Yang, Y.; Duan, N.; and Chen, W. 2024.
\newblock CRITIC: Large Language Models Can Self-Correct With Tool-Interactive
  Critiquing.
\newblock In \emph{Proceedings of International Conference on Learning
  Representations}.

\bibitem[{Guan et~al.(2025)Guan, Zhang, Liu, Shang, Sun, Zhu, Yang, and
  Yang}]{guanrstar}
Guan, X.; Zhang, L.~L.; Liu, Y.; Shang, N.; Sun, Y.; Zhu, Y.; Yang, F.; and
  Yang, M. 2025.
\newblock rStar-Math: Small LLMs Can Master Math Reasoning With Self-Evolved
  Deep Thinking.
\newblock In \emph{Proceedings Of International Conference On Machine
  Learning}.

\bibitem[{Gulati et~al.(2025)Gulati, Miranda, Chen, Xia, Fronsdal, Dumont, and
  Koyejo}]{gulati2025putnamaxiom}
Gulati, A.; Miranda, B.; Chen, E.; Xia, E.; Fronsdal, K.; Dumont, B. d.~M.; and
  Koyejo, S. 2025.
\newblock Putnam-AXIOM: A Functional And Static Benchmark For Measuring Higher
  Level Mathematical Reasoning In LLMs.
\newblock In \emph{Proceedings Of International Conference On Machine
  Learning}.

\bibitem[{Hendrycks et~al.(2021)Hendrycks, Burns, Kadavath, Arora, Basart,
  Tang, Song, and Steinhardt}]{hendrycks2021measuring}
Hendrycks, D.; Burns, C.; Kadavath, S.; Arora, A.; Basart, S.; Tang, E.; Song,
  D.; and Steinhardt, J. 2021.
\newblock Measuring Mathematical Problem Solving With The Math Dataset.
\newblock \emph{arXiv Preprint arXiv:2103.03874}.

\bibitem[{Huang et~al.(2024)Huang, Chen, Mishra, Zheng, Yu, Song, and
  Zhou}]{ICLR2024_8b4add8b}
Huang, J.; Chen, X.; Mishra, S.; Zheng, H.~S.; Yu, A.; Song, X.; and Zhou, D.
  2024.
\newblock Large Language Models Cannot Self-Correct Reasoning Yet.
\newblock In \emph{Proceedings of the International Conference on Learning
  Representations}.

\bibitem[{Kirchner et~al.(2024)Kirchner, Chen, Edwards, Leike, McAleese, and
  Burda}]{kirchner2024prover}
Kirchner, J.~H.; Chen, Y.; Edwards, H.; Leike, J.; McAleese, N.; and Burda, Y.
  2024.
\newblock Prover-Verifier Games Improve Legibility Of Llm Outputs.
\newblock \emph{arXiv Preprint arXiv:2407.13692}.

\bibitem[{Li et~al.(2024)Li, Beeching, Tunstall, Lipkin, Soletskyi, Huang,
  Rasul, Yu, Jiang, Shen et~al.}]{li2024numinamath}
Li, J.; Beeching, E.; Tunstall, L.; Lipkin, B.; Soletskyi, R.; Huang, S.;
  Rasul, K.; Yu, L.; Jiang, A.~Q.; Shen, Z.; et~al. 2024.
\newblock Numinamath: The Largest Public Dataset In Ai4maths With 860K Pairs Of
  Competition Math Problems And Solutions.
\newblock \emph{Hugging Face Repository}, 13(9): 9.

\bibitem[{Liang et~al.(2025)Liang, Qiang, Li, He, Guo, Zhu, Zhang, and
  Cui}]{liang2025mathclean}
Liang, H.; Qiang, M.; Li, Y.; He, Z.; Guo, Y.; Zhu, Z.; Zhang, W.; and Cui, B.
  2025.
\newblock Mathclean: A Benchmark For Synthetic Mathematical Data Cleaning.
\newblock \emph{arXiv Preprint arXiv:2502.19058}.

\bibitem[{Lin et~al.(2024)Lin, Gou, Liang, Luo, Liu, and
  Yang}]{lin2024criticbench}
Lin, Z.; Gou, Z.; Liang, T.; Luo, R.; Liu, H.; and Yang, Y. 2024.
\newblock CriticBench: Benchmarking LLMs For Critique-Correct Reasoning.
\newblock In \emph{Proceedings of Association For Computational Linguistics},
  1552--1587.

\bibitem[{Liu et~al.(2025)Liu, Zeng, Xiao, He, Liu, Wang, Yan, Shen, Zhang, Xu
  et~al.}]{liu2025skywork}
Liu, C.~Y.; Zeng, L.; Xiao, Y.; He, J.; Liu, J.; Wang, C.; Yan, R.; Shen, W.;
  Zhang, F.; Xu, J.; et~al. 2025.
\newblock Skywork-Reward-V2: Scaling Preference Data Curation Via Human-Ai
  Synergy.
\newblock \emph{arXiv Preprint arXiv:2507.01352}.

\bibitem[{Maity and Saikia(2025)}]{maity2025large}
Maity, S.; and Saikia, M.~J. 2025.
\newblock Large Language Models In Healthcare And Medical Applications: A
  Review.
\newblock \emph{Bioengineering}, 12(6): 631.

\bibitem[{Ouyang et~al.(2022)Ouyang, Wu, Jiang, Almeida, Wainwright, Mishkin,
  Zhang, Agarwal, Slama, Ray et~al.}]{ouyang2022training}
Ouyang, L.; Wu, J.; Jiang, X.; Almeida, D.; Wainwright, C.; Mishkin, P.; Zhang,
  C.; Agarwal, S.; Slama, K.; Ray, A.; et~al. 2022.
\newblock Training Language Models To Follow Instructions With Human Feedback.
\newblock \emph{Advances In Neural Information Processing Systems}, 35:
  27730--27744.

\bibitem[{Rafailov et~al.(2023)Rafailov, Sharma, Mitchell, Manning, Ermon, and
  Finn}]{rafailov2023direct}
Rafailov, R.; Sharma, A.; Mitchell, E.; Manning, C.~D.; Ermon, S.; and Finn, C.
  2023.
\newblock Direct Preference Optimization: Your Language Model Is Secretly A
  Reward Model.
\newblock \emph{Advances In Neural Information Processing Systems}, 36:
  53728--53741.

\bibitem[{Schulman et~al.(2017)Schulman, Wolski, Dhariwal, Radford, and
  Klimov}]{schulman2017proximal}
Schulman, J.; Wolski, F.; Dhariwal, P.; Radford, A.; and Klimov, O. 2017.
\newblock Proximal Policy Optimization Algorithms.
\newblock \emph{arXiv Preprint arXiv:1707.06347}.

\bibitem[{Shalev-Shwartz, Shammah, and Shashua(2017)}]{shalev2017formal}
Shalev-Shwartz, S.; Shammah, S.; and Shashua, A. 2017.
\newblock On A Formal Model Of Safe And Scalable Self-Driving Cars.
\newblock \emph{arXiv Preprint arXiv:1708.06374}.

\bibitem[{Shao et~al.(2024)Shao, Wang, Zhu, Xu, Song, Bi, Zhang, Zhang, Li
  et~al.}]{shao2024deepseekmath}
Shao, Z.; Wang, P.; Zhu, Q.; Xu, R.; Song, J.; Bi, X.; Zhang, H.; Zhang, M.;
  Li, Y.; et~al. 2024.
\newblock Deepseekmath: Pushing The Limits Of Mathematical Reasoning In Open
  Language Models.
\newblock \emph{arXiv Preprint arXiv:2402.03300}.

\bibitem[{Tyen et~al.(2024)Tyen, Mansoor, C{\u{a}}rbune, Chen, and
  Mak}]{tyen2024llms}
Tyen, G.; Mansoor, H.; C{\u{a}}rbune, V.; Chen, Y.~P.; and Mak, T. 2024.
\newblock LLMs Cannot Find Reasoning Errors, But Can Correct Them Given The
  Error Location.
\newblock In \emph{Proceedings of Association For Computational Linguistics},
  13894--13908.

\bibitem[{Wang et~al.(2023)Wang, Lyu, Ji, Zhang, Yu, Shi, and
  Tu}]{wang2023document}
Wang, L.; Lyu, C.; Ji, T.; Zhang, Z.; Yu, D.; Shi, S.; and Tu, Z. 2023.
\newblock Document-Level Machine Translation With Large Language Models.
\newblock In \emph{Proceedings of Empirical Methods In Natural Language
  Processing}, 16646--16661.

\bibitem[{Wang et~al.(2024)Wang, Xu, Li, Zhang, Liang, Tang, Yu, and
  Wen}]{wang2024large}
Wang, S.; Xu, T.; Li, H.; Zhang, C.; Liang, J.; Tang, J.; Yu, P.~S.; and Wen,
  Q. 2024.
\newblock Large Language Models For Education: A Survey And Outlook.
\newblock \emph{arXiv Preprint arXiv:2403.18105}.

\bibitem[{Wei et~al.(2022)Wei, Wang, Schuurmans, Bosma, Xia, Chi, Le, Zhou
  et~al.}]{wei2022chain}
Wei, J.; Wang, X.; Schuurmans, D.; Bosma, M.; Xia, F.; Chi, E.; Le, Q.~V.;
  Zhou, D.; et~al. 2022.
\newblock Chain-Of-Thought Prompting Elicits Reasoning In Large Language
  Models.
\newblock \emph{Advances In Neural Information Processing Systems}, 35:
  24824--24837.

\bibitem[{Ying et~al.(2024)Ying, Liu, Liu, and Tao}]{ying2024unveiling}
Ying, Z.; Liu, A.; Liu, X.; and Tao, D. 2024.
\newblock Unveiling The Safety Of Gpt-4o: An Empirical Study Using Jailbreak
  Attacks.
\newblock \emph{arXiv Preprint arXiv:2406.06302}.

\bibitem[{Zhang et~al.(2025)Zhang, Li, Long, Zhang, Lin, Yang, Xie, Yang, Liu,
  Lin et~al.}]{zhang2025qwen3}
Zhang, Y.; Li, M.; Long, D.; Zhang, X.; Lin, H.; Yang, B.; Xie, P.; Yang, A.;
  Liu, D.; Lin, J.; et~al. 2025.
\newblock Qwen3 Embedding: Advancing Text Embedding And Reranking Through
  Foundation Models.
\newblock \emph{arXiv Preprint arXiv:2506.05176}.

\bibitem[{Zhu, Yang, and Sun(2024)}]{zhu2024halueval}
Zhu, Z.; Yang, Y.; and Sun, Z. 2024.
\newblock Halueval-Wild: Evaluating Hallucinations Of Language Models In The
  Wild.
\newblock \emph{arXiv Preprint arXiv:2403.04307}.

\end{thebibliography}

\end{document}